\documentclass[conference]{IEEEtran}
\IEEEoverridecommandlockouts
\usepackage{cite}
\usepackage{amsmath,amssymb,amsfonts}
\usepackage{algorithmic}
\usepackage{graphicx}
\usepackage{textcomp}
\usepackage{xcolor}

\usepackage{booktabs}

\usepackage{multicol}
\usepackage{multirow}

\usepackage{pifont}
\newcommand{\cmark}{\ding{51}}%
\newcommand{\xmark}{\ding{55}}%

\def\BibTeX{{\rm B\kern-.05em{\sc i\kern-.025em b}\kern-.08em
    T\kern-.1667em\lower.7ex\hbox{E}\kern-.125emX}}
\begin{document}

\title{Progressive Multi-Modality Learning for Inverse Protein Folding\\
}

\author{\IEEEauthorblockN{1\textsuperscript{st} \textbf{Jiangbin Zheng}}
\IEEEauthorblockA{\textit{Zhejiang University} \\
\textit{AI Lab, Westlake University}\\
Hangzhou, China \\
zhengjiangbin@westlake.edu.cn}
\and
\IEEEauthorblockN{2\textsuperscript{nd} \textbf{Stan Z. Li}}
\IEEEauthorblockA{\textit{AI Lab, Westlake University}\\
Hangzhou, China \\
Stan.ZQ.Li@westlake.edu.cn}
}

\maketitle

\begin{abstract}
While deep generative models show promise for learning inverse protein folding directly from data, the lack of publicly available structure-sequence pairings limits their generalization. Previous improvements and data augmentation efforts to overcome this bottleneck have been insufficient. To further address this challenge, we propose a novel protein design paradigm called MMDesign, which leverages multi-modality transfer learning. To our knowledge, MMDesign is the first framework that combines a pretrained structural module with a pretrained contextual module, using an auto-encoder (AE) based language model to incorporate prior protein semantic knowledge. Experimental results, only training with the small dataset, demonstrate that MMDesign consistently outperforms baselines on various public benchmarks. To further assess the biological plausibility, we present systematic quantitative analysis techniques that provide interpretability and reveal more about the laws of protein design. 
\end{abstract}


\begin{figure*}[t]
    \centering
    \includegraphics[width=0.85\linewidth]{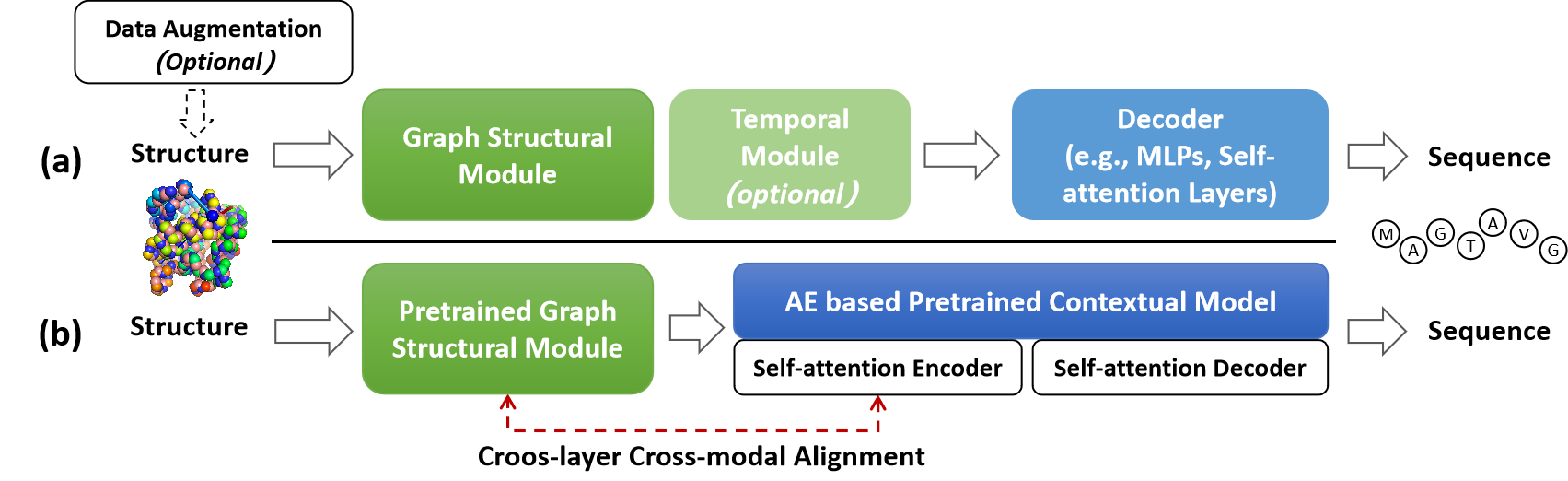}
    \caption{(a) The mainstream advanced deep protein design framework, where the GNN-based structure module represents structural features, followed by an optional contextual module, and the decoder responsible for decoding sequences. (b) The proposed novel paradigm of protein design framework consists of two pretrained modules, i.e., the pretrained structure module and the auto-encoder (AE) contextual module. Additional cross-layer cross-modal alignment aims to enhance constraints.}
    \label{fig:1}
\end{figure*}

\section{Introduction}
\label{sec:intro}
De novo protein design, also known as protein inverse folding, aims to generate protein sequences based on corresponding protein coordinates. Protein design is crucial for various applications in bioengineering, including drug design and enzyme discovery, but remains challenging.
Recent years have seen significant progress in protein design with the emergence of deep learning. In particular, supervised generative models have enabled more efficient and effective prediction of protein sequences\cite{ingraham2019generative,strokach2020fast,anand2022protein,jing2020learning,hsu2022learning,dauparas2022robust,zhao2023protein,zheng2023mmdesign}. Deep generative models show the potential to learn protein design rules directly from data, making them a promising alternative to traditional approaches.

The mainstream and advanced generative protein design frameworks typically follow an encoder-decoder architecture, as illustrated in Figure~\ref{fig:1}(a). A graph is used to represent the structural information in Euclidean space, and graph neural networks (GNNs) are responsible for encoding geometric information into high-dimensional space. Then the decoder recovery the protein sequences through either autoregressive networks based on self-attention\cite{ingraham2019generative,jing2020learning} or non-autoregressive networks\cite{gao2022alphadesign,gao2022pifold,zheng2023lightweight}.
However, the lack of massively available parallel datasets has become a major bottleneck, leading to problems such as under-training or over-fitting. Although billions of protein sequences are available, few protein structures have been experimentally determined. Hence, it is still challenging to train a practical protein design model compared to established cross-modal tasks such as vision-language works.

To address this limitation, various solutions have emerged. 1) The most common method is to focus on protein design architecture, like enhancing the structural equivalent representation, as shown in Figure~\ref{fig:1}(a). But at this stage of development, this will only marginally improve performance. 2) The most straightforward approach is to expand the dataset by generating large-scale pseudo-structures\cite{hsu2022learning} sampled from advanced protein structure prediction models. However, the data construction process is tedious and time-consuming. 3) Another newly emerging framework is to introduce a pretrained structural module, which greatly reduces the reliance on supervised data while maintaining satisfactory performance \cite{zheng2023lightweight}. 
Among these solutions, utilizing pretrained structural networks has become a consensus. To further enhance semantic knowledge, it is worth considering injecting the prior language knowledge of sequences since there are large-scale sequence resources. However, incorporating prior knowledge of both source (structural) and target (textual) modalities is intractable for supervised modes, particularly in protein design where there are no reference works.

Fortunately, a new paradigm in the visual-textual domain \cite{zheng2023cvt} enlightens us to propose a novel paradigm for generative protein design, MMDesign, which leverages multi-modality transfer learning through pretrained structural and contextual modules, as illustrated in Figure~\ref{fig:1}(b). This framework fully exploits pretraining models to overcome the data bottleneck. For the structural module, we incorporate an advanced equivariant network. For the contextual module, we introduce an auto-encoder (AE)-like manner, which can integrate comprehensive prior knowledge and assume the functions of encoding and decoding. 
Furthermore, we incorporate explicit cross-modal alignment constraints\cite{min2021visual,hao2021self,zheng2022using,zhang2024learnable}. This alignment can be viewed as a \textit{consistency} to measure the distance between different distributions \cite{zuo2022c2slr,zheng2021enhancing,zheng2023cvt,lou2023refining,yao2023ndc,Zhaowfdiff}. Maintaining consistency across modalities allows the structural module to be constrained by long-range temporal context, as shown in Figure~\ref{fig:1}(b). The AE module here serves similar functions as \textit{consistency}.

MMDesign is trained solely on the small dataset CATH \cite{orengo1997cath,ingraham2019generative}, yet extensive experiments demonstrate that it substantially outperforms the state-of-the-art baselines without relying on large-scale supervised data. 
However, previous studies have focused primarily on average performance, with little attention paid to explaining the generated results and their biological plausibility. We present the first systematic quantitative and comprehensive analysis techniques to assess the plausibility of the generated protein sequences and the distribution of data. 

The main contributions are as follows:

$\bullet$ The proposed MMDesign innovatively introduces both prior structural and contextual semantic knowledge based on pretraining. This provides a new way of thinking to overcome the bottleneck of other cross-modal protein tasks.

$\bullet$ The introduction of explicit cross-modal alignment between the structural and contextual modality using a cross-layer consistency constraint and implicit cross-modal alignment using favourable properties of the AE mechanism.

$\bullet$ Despite only training on a small dataset, MMDesign substantially outperforms baselines and even surpasses the strong baselines trained on large datasets.
More comprehensive interpretable analyses are presented for multidimensional validation and reveal the underlying protein design patterns.

\begin{figure*}[htp]
    \centering
    \includegraphics[width=0.85\linewidth]{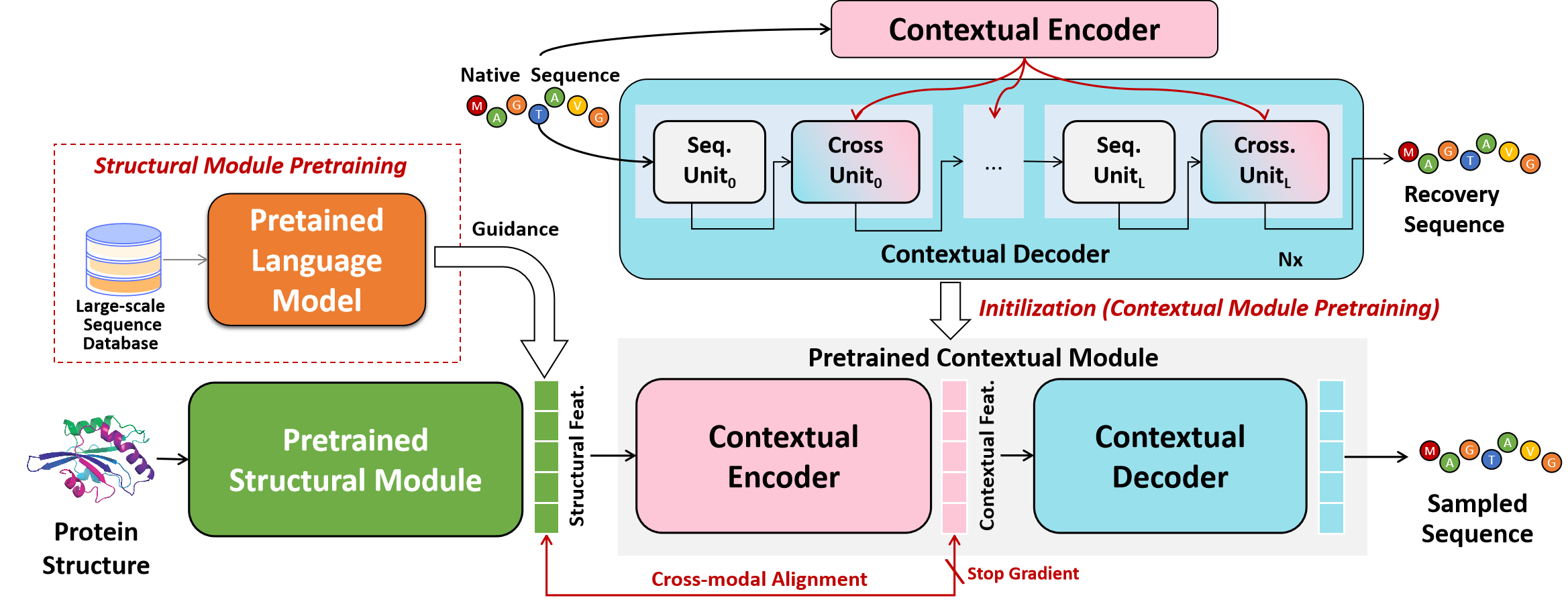}
    \caption{Diagram of the MMDesign framework. The proposed training pipeline is divided into two steps: Step 1 involves pretraining the structural module and the contextual module separately, while Step 2 entails the pretrained modules transferred from Step 1 to optimize the overall MMDesign framework.}
    \label{fig:2}
\end{figure*}

\section{Related Work}
\label{sec:relatedwork}
Recently, generative models for inverse protein folding have emerged gradually. For instance, Structured Transformer\cite{ingraham2019generative} introduced the formulation of fixed-backbone design as a conditional sequence generation problem and used invariant features with GNNs. GVP-GNN \cite{jing2020learning} further improved GNNs with translation- and rotation-equivariance to enable geometric reasoning, highlighting the powerful structural geometric vector perceptron (GVP).
More recently, ProteinMPNN \cite{dauparas2022robust} achieved higher-level sequence recovery and improved inference efficiency.
Then PiFold \cite{gao2022pifold} contained a novel residue featurizer and GNN layers to predict protein sequences in a one-shot way. But it struggles with longer sequences.
ESM-IF\cite{hsu2022learning} utilized GVP-based encoder to extract geometric features followed by a Transformer. It also employed a data augmentation strategy with nearly 12 million pseudo-protein structures, achieving an impressive native sequence recovery.
However, scaling up the data by mostly three orders of magnitude significantly increases the training difficulty. Moreover, its better performance does not extend to single and short chains.

\section{Methods}
\label{sec:methods}
As depicted in Figure~\ref{fig:2}, MMDesign framework consists of a structural module and a contextual module sequentially.
The goal of the MMDesign is to generate amino acid sequences based on the atomic coordinates (N, C$_\alpha$, C) of the corresponding  protein backbones.
The overall training pipeline is divided into two main steps. Step 1 is to initialize the GVPConv-based structural module and pretrain the AE network for the contextual module. The parameters of the structural module are directly derived from the off-the-shelf protein structure model \cite{hsu2022learning}. And step 2 is to transfer the pretrained structural module and contextual module into the MMDesign framework. 

\subsection{Pretraining Stage}

\textbf{Geometric Structural Module Initialization}.
The GVPConv structural module is a key component. 
GVP has been widely used in protein design\cite{jing2020learning,hsu2022learning} due to its lightweight and efficient linear transformation, as well as its ability to learn rotation-equivariant transformations of vector features and rotation-invariant transformations of scalar features. In our framework, we use a variant GVP as in ESM-IF \cite{hsu2022learning}, which transforms the vector features into local reference frames defined for each amino acid to derive rotation-invariant features.
The full graph convolution following GVP utilizes message passing to update the node embeddings in each propagation step, taking into account messages from the neighbor nodes and edges \cite{hsu2022learning,jing2020learning}.
To obtain a better representation of the protein structures, we initialize the GVPConv parameters as \cite{zheng2023lightweight}, where the GVPConv module is pretrained with guidance from a large-scale protein language model through contrastive alignment constraints.

\textbf{Contextual Transformer Pretraining}.
Our AE model is a generic autoregressive encoder-decoder Transformer architecture\cite{vaswani2017attention}. The only modification is the use of learned positional embeddings instead of classical sinusoidal positional embeddings. 
Shared token embedding and shared positional encoding are used since the encoder and decoder share the same amino acid tokens.
To pretrain the AE solely based on the sequence data from the CATH training set, we employ a sequence-to-sequence recovery task as a pseudo-machine translation process, as shown in Figure~\ref{fig:2}. This enables the AE to learn contextual semantic knowledge. And cross-entropy loss is used to learn recovering sequences. 
Formally, we assume that the input protein sequence of the encoder is denoted as $S_\text{in} = \{s_1, s_2, \cdots, s_n\}$ with $n$ amino acids, and the generated sequence from decoder is denoted as $S_\text{out} = \{s'_1, s'_2, \cdots, s'_n\}$, which corresponds to the probability distribution denoted as $S_\text{logits}$. The reference native sequence is $S_\text{native} = S_\text{in} = \{s_1, s_2, \cdots, s_n\}$. The cross-entropy loss is as follows:
\begin{equation}
\mathcal{L}_{\textrm{seqCE}} = \textrm{CE} \big(\textrm{Softmax} (S_\textrm{logits}), S_\textrm{native} \big).
\end{equation}


\subsection{Protein Design Framework Training}

\textbf{MMDesign Training Pipeline}.
The MMDesign framework concatenates the structural module and the contextual module sequentially, as illustrated in the lower section of Figure~\ref{fig:2}. The contextual module only preserves the encoder and decoder layers of the AE model in step 1 and discards the embedding layers.
Let $X=\{x_1, x_2, \cdots, x_n \}$ denote the atomic coordinates of the protein backbone of a given structure-sequence pair, where $n$ is the length of the protein sequence, and $x_i \in \mathbb{R}^{3 \times 3}$ represents the $i-$th 3D coordinates of the amino acid residue, consisting of N, C$_{\alpha}$, and C atoms. The corresponding generated protein sequence is represented as $Y = \{Y_1, Y_2, \cdots, Y_n \} \in \mathbb{R}^{n}$ with the same length as $X$, where $Y_i$ denotes the $i-$th amino acid residue in textual form. The reference native sequence of $Y$ is $\hat{Y} = \{ \hat{Y}_1, \hat{Y}_2, \cdots, \hat{Y}_n \} \in \mathbb{R}^{n}$.
Firstly, the protein backbone $X$ is input to the GVPConv structural module to obtain the geometric structural features $Z_\text{struc}= \{ z_0, z_1, \cdots, z_n \}$ with a dimension $d$.
Then, the structural features $Z_\text{struc}$ are fed to the encoder-decoder layers of the contextual module sequentially. The features obtained by the encoder are referred to contextual features $Z_\text{seq}=\{z'_0, z'_1, \cdots, z'_n \}$. Finally, the decoder generate the sequence distribution $D_{\text{logits}} =\{o_0, o_1, \cdots, o_n \}$ as:
\begin{equation}
Z_\text{struc} = \text{GVP}(X) \in \mathbb{R}^{n \times d},
\end{equation}
\begin{equation}
Z_\text{seq} = \textrm{Encoder}(Z_\text{struc}) \in \mathbb{R}^{n \times d}.
\end{equation}
\begin{equation}
\begin{split}
    D_{\text{logits}} & = \textrm{Decoder}(Z_\text{seq}) \in \mathbb{R}^{n \times M},\\
    Y & = \text{argmax}\big( \text{Softmax}(D_{\text{logits}})\big) \in \mathbb{R}^{n},
\end{split}
\end{equation}
where $M$=20 denotes 20 types of amino acids.

\begin{table*}[htp]
\centering
\scalebox{0.85}{
\begin{tabular}{clcccccccccc} 
    \toprule

    \multirow{2}{*}{\textbf{Group}} & \multirow{2}{*}{\textbf{Models}} & \multicolumn{5}{c}{\textbf{Perplexity}} & \multicolumn{5}{c}{\textbf{Recovery (\%)}} \\ 

    \cmidrule(lr){3-7}\cmidrule(lr){8-12}
    
    &  & \textbf{All} & \textbf{Short} & \textbf{Single-chain}  & \textbf{Ts50} & \textbf{Ts500} & \textbf{All}& \textbf{Short} & \textbf{Single-chain}  & \textbf{Ts50} & \textbf{Ts500} \\ 
    \midrule
 
    \multirow{12}{*}{1} 
    & *Natural frequencies \cite{hsu2022learning} & 17.97   & 18.12 & 18.03  & - & - & 9.5 & 9.6 & 9.0 & - & -\\
    & SPIN \cite{li2014direct} & - & -  & - & - & - & - & - & - & 30.3 & 30.3 \\
    & SPIN2 \cite{o2018spin2} & - & 12.11 & 12.61  & - & - & - & - &- & 33.6 &36.6 \\
    & Structured Transformer \cite{ingraham2019generative} & 6.85 & 8.54 & 9.03  & 5.60 & 5.16 & 36.4 & 28.3 & 27.6 & 42.40 &44.66 \\
    & Structured GNN \cite{jing2020learning} & 6.55 & 8.31	& 8.88   & 5.40 & 4.98 & 37.3 & 28.4 & 28.1 & 43.89 & 45.69\\
    
    & *ESM-IF \cite{hsu2022learning} & 6.44 &  8.18 & 6.33  & - & - & 38.3  & 31.3 & 38.5 & - &-\\
    
    & GCV \cite{tan2022generative} & 6.05 & 7.09 & 7.49  & 5.09 & 4.72 & 37.64 & 32.62 & 31.10  & 47.02 & 47.74\\

    & *GVP-GNN-large \cite{jing2020learning} & 6.17 &  7.68	 & \underline{6.12}  & - & - & 39.2 & 32.6 & \textbf{39.4} & - &-\\
    & GVP-GNN \cite{jing2020learning} & 5.29 & 7.10 & 7.44  & 4.71 & 4.20 & 40.2 & 32.1 & 32.0 & 44.14 & 49.14 \\
    
    & AlphaDesign \cite{gao2022alphadesign} & 6.30 & 7.32 & 7.63  & 5.25 & 4.93 & 41.31 & 34.16 & 32.66  & 48.36 & 49.23\\
    & ProteinMPNN \cite{dauparas2022robust} & 4.61 &  6.21	  & 6.68  & 3.93 & 3.53 & 45.96 & 36.35 & 34.43 & 54.43 & 58.08\\
    
    & PiFold \cite{gao2022pifold} & 4.55 & \underline{6.04} & 6.31  & 3.86 & 3.44 & 51.66 & \underline{39.84} & 38.53 & \underline{58.72} & 60.42\\
    \midrule
    
    \multirow{1}{*}{2}
    & ESM-IF \cite{hsu2022learning} & \underline{3.99} &  6.30	 & 6.29  & \underline{3.43} & \underline{3.34} & \underline{52.51} & 34.74 & 34.25  & 56.66 & \underline{60.85}\\
    \midrule
    
    \multirow{1}{*}{3}
    & Ours (MMDesign) & \textbf{3.86} & \textbf{5.32} & \textbf{5.43}  & \textbf{2.89} & \textbf{3.20} & \textbf{54.88} & \textbf{39.89} & \underline{39.24} & \textbf{59.87} & \textbf{61.06}\\
	 \bottomrule
  
\end{tabular}
}
\caption{Perplexity and recovery comparison. The best results are \textbf{bolded}, and the second best \underline{underlined}. Group 1 \& 3 are trained on a small-scale CATH training set only, while Group 2 is trained on a large-scale CATH+AlaphaDB training set. The \textit{Short} and \textit{Single-chain} are subsets of CATH test set \textit{ALL}. *: evaluated on CATH 4.3; Others are evaluated on CATH 4.2.}
\label{tab:1}
\end{table*}

\textbf{Generation Objective}.
The cross-entropy loss is commonly used in protein design tasks. By minimizing the cross entropy, we can obtain an approximate distribution that better matches the target distribution.
We observed that emphasizing the salient features of the output distribution can further improve the performance of our model. To achieve this, we propose an exponential cross entropy (expCE) that accentuates the spiking effect of the output distribution. The expCE is calculated as follows:

\begin{equation}
\mathcal{L}_\text{expCE} = \exp\left(\sum_{b=0}^{B}\text{CE}(\text{Softmax}(D_{\text{logits}}^{(b)}), \hat{Y}^{(b)})\right),
\end{equation}
where $exp$ denotes the exponential calculation, and $b$ denotes the $b$-th element in a batch of $B$ elements.

\textbf{Cross-modal Alignment}.
We view protein design as a cross-modal task, where the structural features $Z_\text{struc}$ obtained from the GVPConv module and the contextual features $Z_\text{seq}$ obtained from the encoder of the contextual module are considered as two different modalities. To enforce alignment constraints across modalities, we propose a cross-layer cross-modal alignment constraint (CAC) loss, as shown in Figure~\ref{fig:2}. The CAC loss is implemented as a knowledge distillation loss (KL-divergence here), where the entire deep contextual features and the structural features are treated as teacher and student models, respectively. A high temperature $T=8$ is adopted to `soften' the probability distribution from potential spike responses. The distillation process is expressed as:
\begin{equation}
\mathcal{L}_\text{CAC} = \text{KL}(\text{Softmax}(\frac{Z_\text{struc}}{T}), \text{Softmax}(\frac{Z_\text{seq}}{T}) ).
\end{equation}

\textbf{Overall}, $\mathcal{L}_\text{expCE}$ is used to to align the generation distribution with the reference distribution, and the cross-layer cross-modal alignment CAC provides long-range supervision to the structure extractor. The final training objective is composed of the primary $\mathcal{L}_\text{expCE}$ loss and the auxiliary $\mathcal{L}_\text{CAC}$ loss.

\begin{figure*}[t!]
    \centering
    \includegraphics[width=0.9\linewidth]{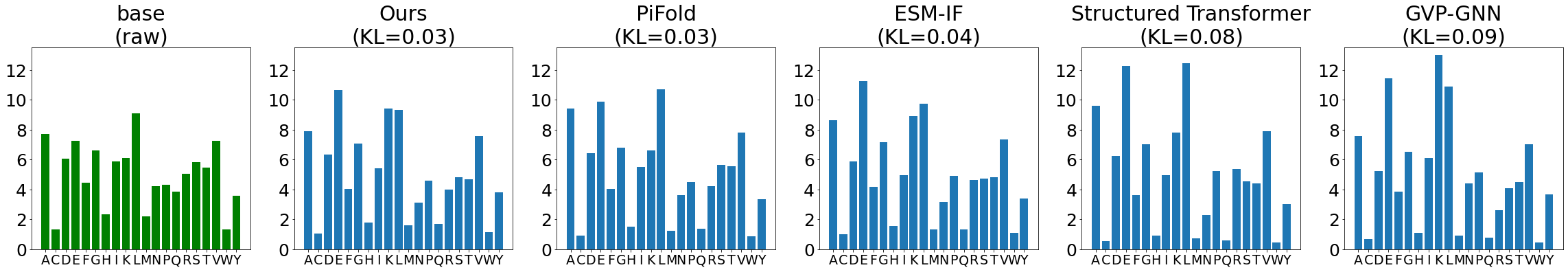}
    \caption{Normalized statistics of the number of amino acid residue types corresponding to the ground-truth sequences (base) and generated sequences of protein design models on CATH test set. The KL value indicates the KL divergence of the generated sequence distribution with respect to the ground-truth sequence distribution.}
    \label{fig:distribution}
\end{figure*}

\section{Experiments}
\subsection{Settings}
\textbf{Datasets}. 
The CATH \cite{ingraham2019generative,orengo1997cath} is the main dataset, in which he training, validation, and testing splits contain 18204, 608, and 1120 structure-sequence pairs, respectively. The training split is used for training MMDesign, and the validation and testing splits are for evaluation purposes. CATH specifically refers to CATH 4.2 if not otherwise specified.
We also report results on Ts50 \& Ts500 \cite{li2014direct} to assess the generalization. 

\textbf{Evaluation Metrics}.
Perplexity and sequence recovery on held-out native sequences are two commonly adopted metrics for protein design tasks. Perplexity measures the inverse likelihood of native sequences in the predicted sequence distribution (lower perplexity for higher likelihood). And sequence recovery measures the accuracy with which the sampled sequence matches the native sequence at each site. 

\textbf{Implementation Details}. 
During training, we employed SGD optimizer with a batch size of 5, a learning rate of \textit{1e-3} to optimize the protein design model, and low temperature \textit{1e-6} for sampling.
The GVPConv consists of 4 layers with a dropout of 0.1, and the top-k neighbors are 30. The node hidden dimensions of scalars and vectors are 1024 and 256, respectively.
For the contextual Transformer, both the encoder and decoder have 8 layers of 8-head multi-head self-attention blocks, with an embedding dimension of 512 and an attention dropout of 0.1.
And 1 NVIDIA A100 80GB GPU is used.

\subsection{Main Results}
In line with previous work\cite{ingraham2019generative,jing2020learning,hsu2022learning,gao2022pifold}, we start with evaluation in the standard setting.

\textbf{Evaluation on CATH Set.}
Table~\ref{tab:1} reports the perplexity (lower is better) and recovery (higher is better) scores on the generic dataset CATH test set (``ALL''). MMDesign achieves the best performances with a perplexity of 3.86 and a recovery of 54.88\%, outperforming the SOTA ProteinMPNN, PiFold (Group 1), and ESM-IF (Group 2) by a significant margin.

\textbf{Evaluation on Single/Short Chains.}
Following \cite{ingraham2019generative}, we also evaluate the performance of the models on the subsets of the CATH test set, i.e., ``Short'' dataset (protein sequence length $\leq$ 100 residues) and ``Single-chain'' dataset (single-chain proteins recorded in the Protein Data Bank), for evaluation, as shown in Table~\ref{tab:1}. MMDesign achieves excellent performances on both ``Single-chain'' and ``Short'' datasets (although the recovery score of ``Single-chain'' is slightly lower than GVP-GNN-large on CATH 4.3). In particular, in terms of perplexity, MMDesign (Short: 5.32; Single-chain: 5.43) significantly outperforms PiFold (Short: 6.04; Single-chain: 6.31) and ProteinMPNN (Short: 6.21; Single-chain: 6.68).

\textbf{Out-of-domain Generalization.}
To fully compare the generalizability of the different models on the out-of-domain datasets, we report the results on the Ts50 and Ts500 datasets in Table~\ref{tab:1}. MMDesign still achieves consistent improvements for Ts50 and Ts500, where the recovery score in Ts50 is close to 60\%, and the score in Ts500 breaks 61\%.

\textbf{Overall}, ESM-IF generalizes better on both generic and out-of-domain datasets, while PiFold and ProteinMPNN perform better on short and single-chain datasets.
However, MMDesign outperforms all baselines across a variety of test sets, including generic datasets, out-of-domain datasets, and short and single-chain datasets with balanced performances, showing strong generalization and robustness. Notably, the results also demonstrate that a pretrained prior knowledge migration-based strategy with small data training can outperform the large data training strategy such as ESM-IF.


\begin{table}
\centering
\scalebox{0.85}{
\begin{tabular}{ccccccccc}\toprule

\multirow{2}{*}{\textbf{\#} }
& \multirow{2}{*}{\textbf{PSM}} & \multirow{2}{*}{\textbf{PCM}} & \multicolumn{2}{c}{\textbf{Perplexity}} & \multicolumn{2}{c}{\textbf{Recovery(\%)}} \\

\cmidrule(lr){4-5} \cmidrule(lr){6-7} 
&  &  & \textbf{Val} & \textbf{Test} & \textbf{Dev} & \textbf{Test}\\ 
\midrule

	1 & \xmark & \xmark & 6.93	& 6.76 &  35.40	& 36.86 \\
	2 & \xmark & \cmark & 6.30  & 6.22 &  39.94	& 41.81 \\
	3 & \cmark & \xmark	& 4.75 	& 4.52 & 47.05	& 49.41 \\
	4 & \cmark & \cmark	& \textbf{4.23} & \textbf{3.86} & \textbf{52.67} & \textbf{54.88}  \\
\bottomrule
  
\end{tabular}
}
\caption{Ablation study of pretrained modules on CATH validation and test set (PSM: pretrained structural module; PCM: pretrained contextual module). The best results are \textbf{bolded}.}
\label{table:3}
\end{table}

\subsection{Performance Analysis}

\textbf{Pretrained Module Ablation}.
To illustrate the significance of the pretrained modules, we conduct an ablation study to evaluate the impact of the pretrained structural module (PSM) and the pretrained contextual module (PCM). Overall, both have enhanced effects on the performance, as shown in Table~\ref{table:3}.
Comparing \#1 vs. \#3 and \#1 vs. \#2, we observe that the pretrained structural features contribute more to the performance. This observation is consistent with our motivation for introducing prior structural knowledge. Moreover, it is likely that the PCM module, as a downstream module of PSM, is directly influenced by the representation capability of PSM.
Surprisingly, comparing only \#1 vs. \#2, performances are improved considerably even without using initialization parameters of the PSM, i.e., by using only the PCM. Interestingly, this is a completely new observation that has not been extensively explored in previous protein design studies. These suggest that the PCM has learned rich knowledge of the protein language to improve protein design tasks. Comparing \#3 with \#4 further confirms this observation.

\textbf{Generated Sequence Distribution}
As shown in Figure~\ref{fig:distribution}, we counted both the generated protein residues and raw protein residues according to amino acid residue categories. We generated approximately 181k amino acid residues for each model based on 1,120 protein sequences from the CATH test set.
Representative protein design models are chosen for comparison with our MMDesign model. To facilitate comparison, we normalized the distribution using the maximum-minimum normalization process.
Compared to the ground-truth distribution (base) derived from raw reference sequences in Figure~\ref{fig:distribution} (green),  our MMDesign and PiFold produce distributions that are visually closer to the base.
To quantitatively characterize the similarity of the generated sequences to the base distribution, KL-divergence is employed. The results are consistent with the visual observations, where our MMDesign and PiFold have lower KL-divergence values (around 0.03) than other baselines, indicating better similarity to the reference distribution. In contrast, GVP-GNN and Structured Transformer perform poorly with higher KL values.

\begin{figure}[t]
    \centering
    \includegraphics[width=0.85\linewidth]{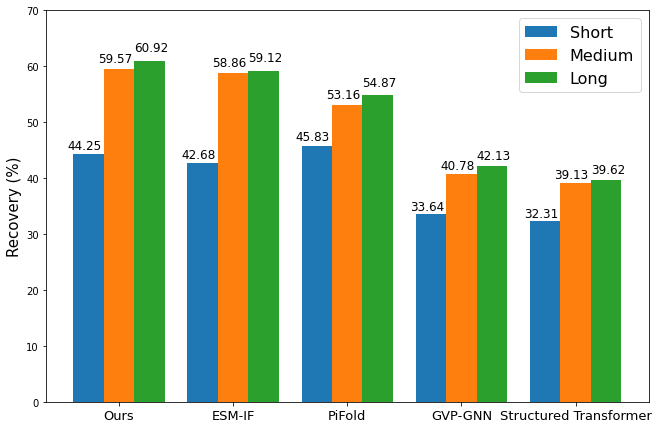}
    \caption{Recovery score comparison of different models on short, medium and long sequences derived from CATH test set (Length division: Short $\in (0,125]$, Medium $\in (125, 188]$, Long $\in (188, 500)$. The three divisions are equal in number).}
    \label{fig:sent_len}
\end{figure}

\textbf{Protein Length Analysis}.
The comparison on the ``Short'' dataset in Table~\ref{tab:1} only initially demonstrates the advantage in short sequences. To further clarify the effect of protein sequence length, we divided the CATH test set into equal numbers of short, medium, and long divisions in order of increasing length (``short'' here is different from ``Short'' in Table~\ref{tab:1}). Figure~\ref{fig:sent_len} illustrates the comparison with existing advanced baselines on these three divisions. 
All models perform much worse on short sequences compared to medium and long sequences, consistent with the previous observation in Table~\ref{tab:1}. In terms of short sequences, MMDesign (44.25\%) performs slightly worse than PiFold (45.83\%) but still outperforms other models greatly. For medium and long sequences, our model has a obvious advantage over others, although ESM-IF does not perform badly either. As expected, the earlier GVP-GNN and Structured Transformer perform worse in all divisions.
Overall, our model is relatively better on short sequences and has a significant performance advantage on longer sentences. The comparison provides some insight into the role of the self-attention mechanism in capturing long-range dependencies for longer sequences (PiFold and GVP-GNN do not use self-attention). 
Although Structured Transformer relies on self-attention for temporal feature representation, it fails to perform well due to the lack of a stronger structural representation module such as GVP.

\section{Conclusion} 
The proposed MMDesign represents a significant step forward in protein design by introducing a novel multi-modality transfer learning and an explicit consistency constraint. 
Trained only on the small dataset, MMDesign improves significantly over baselines across various benchmarks. 
We also systematically present analysis techniques based on biological reasonable perspectives and data distribution statistics, which greatly improve the interpretability and reveal more about the nature and laws of protein design.

\section*{Acknowledgements}
This work was supported by National Science and Technology Major Project (No. 2022ZD0115101), National Natural Science Foundation of China Project (No. U21A20427), Project (No. WU2022A009) from the Center of Synthetic Biology and Integrated Bioengineering of Westlake University and Integrated Bioengineering of Westlake University and Project (No. WU2023C019) from the Westlake University Industries of the Future Research Funding.

\bibliographystyle{IEEEbib}
\bibliography{my_citation}

\end{document}